\def\year{2020}\relax
\documentclass[letterpaper]{article} % DO NOT CHANGE THIS
\usepackage{aaai20}  % DO NOT CHANGE THIS
\usepackage{times}  % DO NOT CHANGE THIS
\usepackage{helvet} % DO NOT CHANGE THIS
\usepackage{courier}  % DO NOT CHANGE THIS
\usepackage[hyphens]{url}  % DO NOT CHANGE THIS
\usepackage{graphicx} % DO NOT CHANGE THIS
\usepackage{multirow}
\usepackage{amsmath}
\usepackage{graphicx}  % DO NOT CHANGE THIS
\title{Spatial-Temporal Self-Attention Network for Flow Prediction }
\author{Haoxing Lin, Weijia Jia, Yiping Sun, Yongjian You\\
State Key Laboratory of Internet of Things for Smart City, FST, University of Macau\\
Shanghai Jiaotong University\\
mb85410@um.edu.mo}
\begin{document}
	\newcommand{\norm}[1]{\left\lVert#1\right\rVert}
	
	\maketitle
	
	\begin{abstract}
		Flow prediction (e.g., crowd flow, traffic flow) with features of spatial-temporal is increasingly investigated in AI research field. It is very challenging due to the complicated spatial dependencies between different locations and dynamic temporal dependencies among different time intervals. Although measurements of both dependencies are employed, existing methods suffer from the following two problems. First, the temporal dependencies are measured either uniformly or bias against long-term dependencies, which overlooks the distinctive impacts of short-term and long-term temporal dependencies. Second, the existing methods capture spatial and temporal dependencies independently, which wrongly assumes that the correlations between these dependencies are weak and ignores the complicated mutual influences between them. To address these issues, we propose a Spatial-Temporal Self-Attention Network (\textbf{ST-SAN}). As the path-length of attending long-term dependency is shorter in the self-attention mechanism, the vanishing of long-term temporal dependencies is prevented. In addition, since our model relies solely on attention mechanisms, the spatial and temporal dependencies can be simultaneously measured. Experimental results on real-world data demonstrate that, in comparison with state-of-the-art methods, our model reduces the root mean square errors by 9\% in inflow prediction and 4\% in outflow prediction on Taxi-NYC data, which is very significant compared to the previous improvement.
	\end{abstract}
	
	\section{Introduction}
	Flow prediction, as one of the most crucial problems in today's smart city research, has drawn increasing attention in AI research field. With a boosted number of population, effective prediction of flow (e.g., crowd flow, traffic flow) becomes more and more critical for first-tier cities. Practically, the performance of various applications, such as intelligent service allocation and dynamic traffic management, benefit from higher prediction accuracies in crowd flow prediction and traffic flow prediction \cite{WuT16}. On the other hand, a more substantial amount of available data has been driving the AI researches on flow prediction as well.
	
	Specifically, flow refers to the number of people or vehicles arriving in (inflow) or departing from (outflow) the observed regions at each time interval. The goal of flow prediction is to predict the flow of future times by deriving spatial-temporal patterns from historical data. Before the era of deep learning, flow prediction has been heavily relying on methods from time series analysis community. Traditional statistic methods such as Auto-Regressive Integrated Moving Average (ARIMA), Kalmen filtering, and Vector Auto-Regressive (VAR) models are widely employed in flow prediction \cite{chandra_2009,Li2012,Moreira-Matias,Shekhar}. Although they are straight-forward and easy to deploy, the incapabilities of traditional methods on measuring complicated spatial dependencies limit their performance.
	
	Recently, deep learning-based methods have shown significant advantages in modeling both spatial and temporal dependencies in flow prediction \cite{Zhang:2017:DSR:3298239.3298479}. However, the existing methods still suffer from incomprehensive measurements of long-term and short-term temporal dependencies. Besides, they also ignore the complicated correlation between the spatial and temporal dependencies as capturing them independently. To be specific, the above problems result from the fundamental structures employed by the current methods. Generally, their structures can be categorized as (1) deep residual convolutional network \cite{ZHANG_TKDE} and (2) convolutional recurrent network \cite{stdn}. Although they all consider both spatial and temporal dependencies, each kind of networks has structural problems that intrinsically limit their performances.
	
	For the deep residual convolutional methods, the spatial dependencies of different time intervals are independently measured by multiple deep residual convolutional neural networks \cite{Kaiming_He_2015}. Without any recurrent structures, they try to handle the temporal dependencies by applying deeper and more nested residual networks. However, as the convolutional results of different time intervals are uniformly measured, this kind of structures overlooks the distinctive impacts of short-term and long-term temporal dependencies.
	
	For those who employ convolutional recurrent structure, they apply recurrent networks such as LSTM \cite{doi:10.1162/neco.1997.9.8.1735} on the convolutional results of different time intervals. However, as the long-term temporal dependencies vanish rapidly via passing through the recurrent networks, it is overwhelmed by the short-term temporal dependencies, which causes the incomprehensive measurement of temporal dependencies. Moreover, the computation of the recurrent structure is very inefficient \cite{NIPS2017_7181}, which deters the convolutional recurrent networks to further improve their performance by applying deeper and more nested structures.
	
	Additionally, both of the structures handle the spatial and temporal dependencies asynchronously, which relies on a false assumption that the correlations between the two factors are weak. However, the assumption ignores the fact that the spatial and temporal dependencies have complicated mutual influences, which is very critical for flow prediction under complex situations.
	
	To overcome these challenges, we propose a \textbf{S}patial-\textbf{T}emporal \textbf{S}elf-\textbf{A}ttention \textbf{N}etwork (\textbf{ST-SAN}), which adopts an innovative spatial-temporal self-attention mechanism. Given its shorter path-length to attend the long-term dependency in the self-attention mechanism, our model avoids the vanishing of long-term temporal dependencies. Besides, since it is merely based on attention mechanisms, ST-SAN captures all dependencies simultaneously and thus are more effective as the spatial and temporal dependencies can interrelate to each other. Moreover, without any recurrent or deep convolutional structures, ST-SAN is very computationally efficient.
	
	The contributions of our work can be summarized as follows:
	
	\begin{itemize}
		\item A spatial-temporal self-attention mechanism is developed to handle sophisticated and dynamic spatial and temporal dependencies simultaneously. To the best of our knowledge, the proposed mechanism is the first method that can measure both dependencies synchronously.
		\item Our model prevents the vanishing of long-term temporal dependencies with the self-attention mechanism, which can attend to both short-term and long-term dependencies through equal-length paths.
		\item A Spatial-Temporal Self-Attention Network is proposed, which is computationally efficient as eschewing recurrent and deep convolutional structures. To the best of our knowledge, ST-SAN is the first deep-learning-based flow prediction methods without both of these two structures.
		\item We evaluate our model on three real-world, large-scale datasets and demonstrate its significant advantages over state-of-the-art baselines.
	\end{itemize}
	
	\section{Related Work}
	\subsection{Deep Learning for Flow Prediction}
	Recently, various works based on deep learning have achieved significant improvement in flow prediction. Firstly, the LSTM \cite{doi:10.1162/neco.1997.9.8.1735} based methods demonstrates excellent performance on capturing temporal dependencies when predicting spatial-temporal flow \cite{DBLP:cui_ke_wang}. Then, convolutional structures were investigated on capturing spatial dependencies in flow prediction tasks \cite{Zhang:2016:DPM:2996913.2997016}. After the deep residual convolutional network is proposed \cite{Kaiming_He_2015}, several works based on deep residual structure achieve significant improvement in capturing spatial-temporal dependencies in flow prediction \cite{zhang_zheng_qi}. Lately, after Convolutional LSTM achieved tremendous success in processing spatial-temporal information \cite{NIPS2015_5955}, several researches employ such convolutional recurrent structure to learn spatial and temporal dependencies and further improve the performance of predicting flow \cite{ke_zheng_yang,Zhou:2018:PMC:3159652.3159682,a98d8116a2684b17bdabc50c1e1713b3,stdn}. However, these works fail to comprehensively measure the temporal dependencies and also overlook the complicated correlations between spatial and temporal dependencies.
	
	\subsection{Self-Attention}
	Recently, self-attention has drawn an enormous amount of attention in natural language processing (NLP). Transformer \cite{NIPS2017_7181}, a fully self-attention framework, has been widely adopted in many state-of-the-art pre-training language models \cite{devlin_2018,radford2019language,xlnet}.
	
	The self-attention mechanism has three advantages over traditional convolutional and recurrent structures. First, impacts of distant series can affect each other's output without passing through recurrent steps, or convolution layers. Second, it can learn long-term dependencies effectively. Third, its layer outputs can be calculated in parallel, which is much faster than a series like the RNN \cite{NIPS2017_7181}. However, we observe that directly applying Transformer on flow prediction does not result in the expected improvement. The possible reason may be that it is initially designed for modeling dependencies among a sequence of words, which inherently lacks the consideration of spatial information.
	
	\begin{figure}[t]
		\centering
		\includegraphics[scale = 0.33]{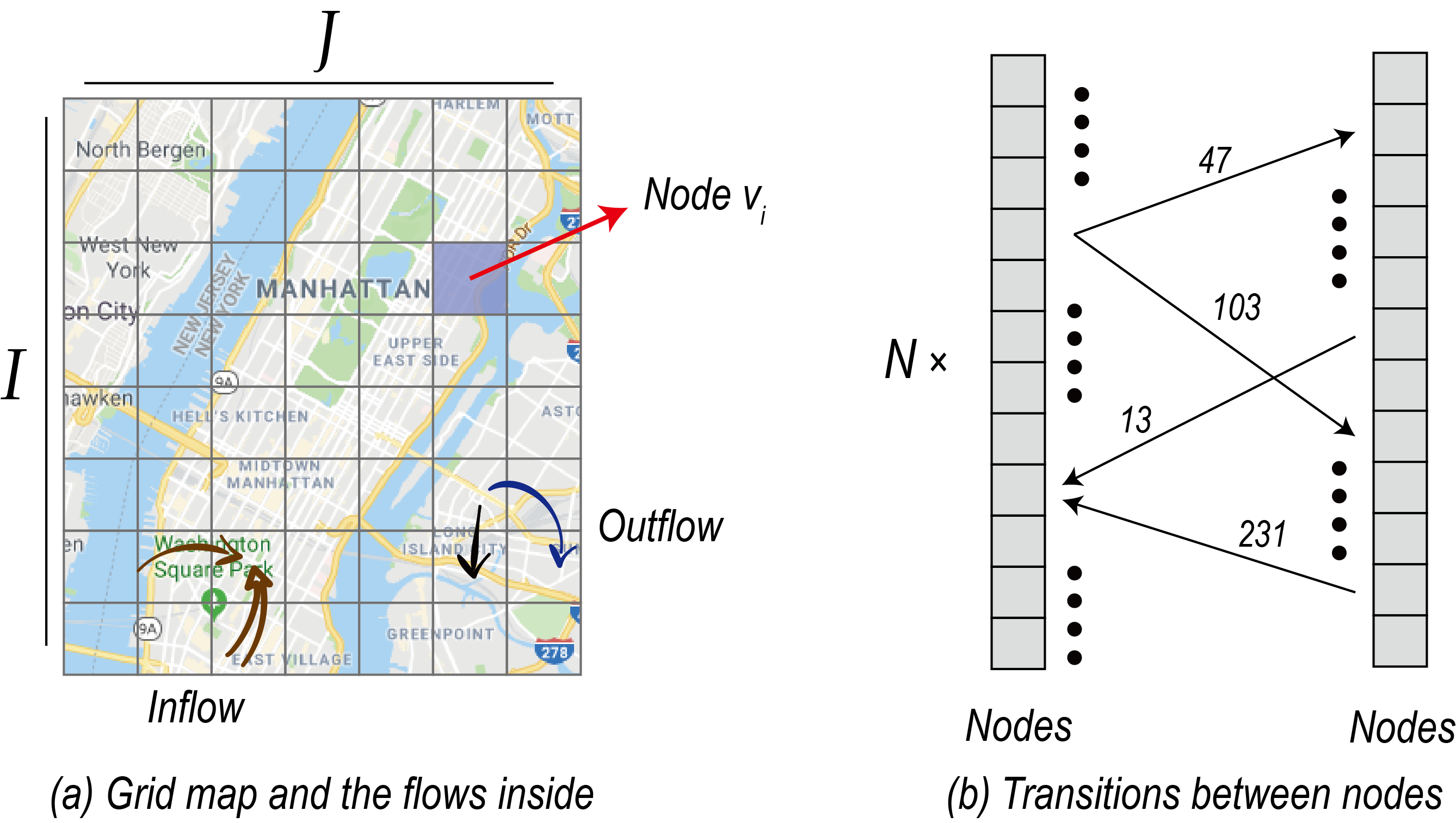} % Reduce the figure size so that it is slightly narrower than the column. Don't use precise values for figure width.This setup will avoid overfull boxes.
		\caption{Map segmentation and the transitions between nodes}
		% \label{fig1}
	\end{figure}
	
	\begin{figure*}[t]
		\centering
		\includegraphics[scale = 0.55]{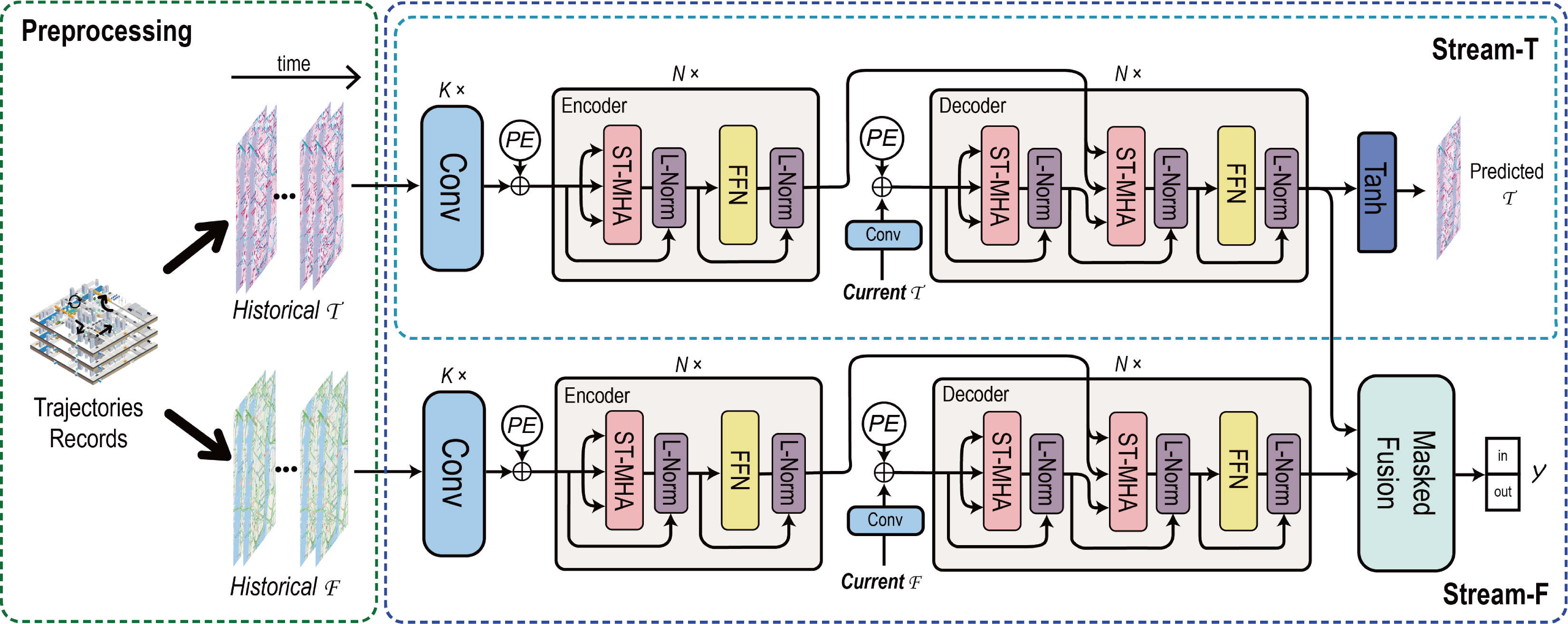} % Reduce the figure size so that it is slightly narrower than the column. Don't use precise values for figure width.This setup will avoid overfull boxes. 
		\caption{Model architecture. PE: positional encoding}
		% \label{fig2}
	\end{figure*}
	
	\section{Notations and Problem Formulation}
	As shown in Figure 1, the spatial area is divided into a $\textit{I} \times \textit{J}$ grid map with \textit{N} grids in total (\textit{N = $I \times J$}). Each grid represents a node (region) in the spatial map, denoted as \{$v_1$, $v_2$, ..., $v_n$\}. \textit{T} stands for the number of all available time intervals equally divided from the whole period. In each time interval, w types of flows (e.g., inflow and outflow) are included in each node, their volumes are determined based on the historical records of object trajectories. Specifically, take inflow and outflow as example, when an object (e.g., person, vehicle) was in $v_s$ at time $t_s$ and appeared in $v_e$ at time $t_e$ ($v_s$ $\neq$ $v_e$, $t_s$ $\leq$ $t_e$), it contributed one volume to each of $v_s$'s outflow and $v_e$'s inflow. The overall volumes of inflow and outflow of $v_i$ at time \textit{t} are denoted as \( \mathcal{F}_{i,t}^{0} \) and \( \mathcal{F}_{i,t}^{1} \). At the meantime, the transitions between nodes are extracted, denoted as \(\mathcal{M}_{i,j}^{t,0}\) for transitions arrive in $v_i$ from $v_j$ and \(\mathcal{M}_{i,j}^{t,1}\) for transitions depart from $v_i$ to $v_j$. Notice that, since the transitions may span across multiple time intervals, we discard those with duration longer than a threshold \textit{m} as they have less effect on flow prediction in the next time interval. After obtaining the historical flow and transition data with length \textit{T} alongside the time axis, we constitute tensors \(\mathcal{F} \in \mathbf{R}^{I \times J \times T \times w}\) and \(\mathcal{M} \in \mathbf{R}^{I \times J \times I \times J \times T \times w}\).
	
	\noindent \textbf{Problem Statement} Given historical flow and transition data \(\mathcal{F}\), \(\mathcal{M}\) as inputs, the task of prediction problem is to learn a function $\textit{f}_\theta$ that maps the inputs to the predicted values $\hat{Y}$ of all nodes at the next time:
	
	\begin{equation}
	\hat{Y} = \textit{f}_\theta(\mathcal{F}, \mathcal{T})
	\end{equation}
	
	\noindent where \(\hat{Y} \in \mathbf{R}^{N \times 2}\) and $\theta$ stands for the learnable parameters.
	
	\section{Model Architecture}
	Figure 2 shows the architecture of ST-SAN, which consists of 2 streams of self-attention networks -- Stream-T and Stream-F. Each of them contains a stack of convolutional layers, an encoder, and a decode. The Stream-T is trained independently on capturing features of transition before merging with Stream-F by a masked fusion mechanism. The detail of each component is described in the following subsections.
	
	\subsection{Encoder and Decoder}
	We employ the encoder-decoder architecture as in most competitive neural sequence transduction models \cite{NIPS2017_7181}. Here, the encoder maps an inputs sequence of historical flow or transition data (\(\mathcal{F}_{hist}\) or \(\mathcal{T}_{hist}\)) to a sequence of continuous representations \textbf{Z}. Given \textbf{Z} and the current flow or transition data (\(\mathcal{F}_{curr}\) or \(\mathcal{T}_{curr}\)), the decoder then generates an output \textit{y} as the predicted output of the next time interval.
	
	The encoder contains a stack of \textit{N} = 4 identical layers, whose sub-layers includes a spatial-temporal multi-head self-attention mechanism and a position-wise fully connected feed-forward network. We also employ the residual connection \cite{Kaiming_He_2015} and layer normalization \cite{ba2016layer} around each of the two sub-layers. To be specific, the output of each sub-layers is \(LayerNorm(x + Sublayer(x))\), where \textit{Sublayer(x)} is the function implemented by the sub-layer itself. The dimension of outputs produced by all sub-layers is set to $d_{model}$ = 64, in order to facilitate the residual connections.
	
	The decoder consists of a stack of \textit{N} = 4 identical layers as well. Besides the two sub-layers in each encoder layer, an additional sub-layer is inserted to performs spatial-temporal multi-head attention over the output of the encoder stack. Also, residual connections followed by layer normalizations are implemented around each sub-layers.
	
	\begin{figure*}[t]
		\centering
		\includegraphics[scale = 0.4]{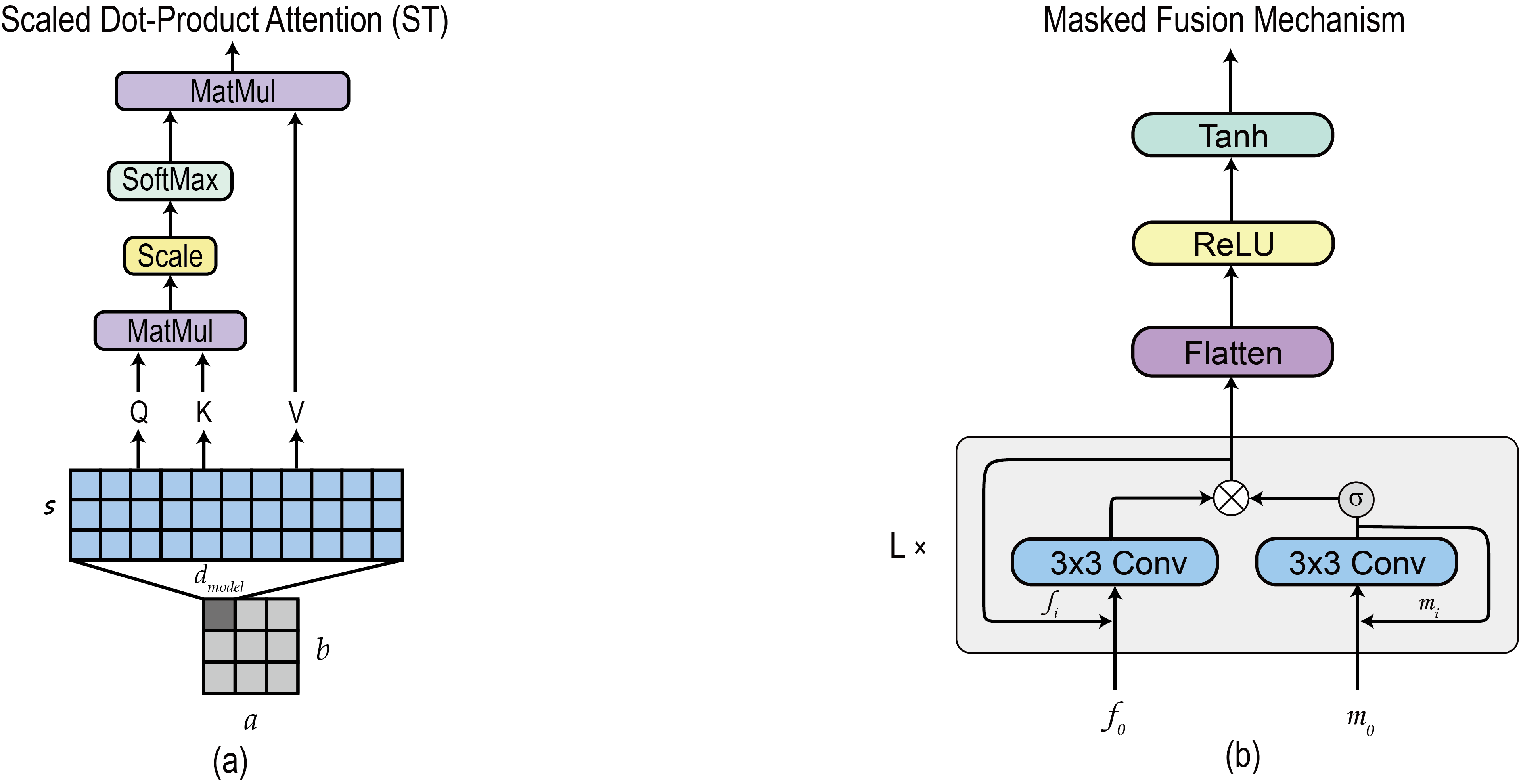} % Reduce the figure size so that it is slightly narrower than the column.
		\caption{The scaled dot-product attention in spatial-temporal self-attention mechanism. (right) Masked Fusion Mechanism. \textbf{$\sigma$} is the \textit{sigmoid} function.}
		% \label{fig3}
	\end{figure*}
	
	\subsection{Spatial-Temporal Self-Attention}
	Compared to ordinary self-attention mechanism adopted in language models, the feature space of the spatial-temporal self-attention mechanism has two more axes inserted to hold the domain of spatial map. As the computation of self-attention can be parallelized \cite{NIPS2017_7181}, an enlarged feature space does not result in longer training time.
	
	In spatial-temporal self-attention, the scaled dot-product attention \cite{NIPS2017_7181} is used as the attention kernel (Figure 3 (a)):
	
	\begin{equation}
	Att(Q,K,V) = softmax(\frac{{QK^T}}{\sqrt{d_k}})V
	\end{equation}
	
	The inputs consist of queries, keys and values, as \textit{Q, K, V} $\in$ \(\mathbf{R^{l \times h \times s \times d_{model}}}\), where $l \times h$ is the size of spatial maps and \textit{h, $d_{model}$} stand for sequence length and feature dimension. The transpose of K is performed between the last 2 axes where \textit{$K^T$} $\in$ \(\mathbf{R^{l \times h \times d_{model} \times s}}\). Also, the matrix multiplication between $\textit{Q}$, $K^T$ is over the last two axes. Then a multi-head attention is constructed upon the scaled dot-product attention:
	
	\begin{equation}
	\begin{aligned}
	ST-&MHA(Q,K,V)\ =\ Concat(h_1,...,h_u)W^O \\
	&where\ h_i = Att(QW_i^Q,KW_i^K,VW_i^V)
	\end{aligned}
	\end{equation}
	
	\noindent where $W_i^Q, W_i^K, W_i^V$ are the learned projection parameter matrices and $u$ is the number of attention head. In this work, we employ u = 8 parallel attention layers, or heads. As the concepts of scaled dot-product attention and multi-head attention have been widely adopted in AI researches, here we exclude their comprehensive descriptions and refer readers to \cite{NIPS2017_7181}.
	
	\subsection{Local Convolution and Area of Interest}
	Before passing the spatial-temporal data into the spatial-temporal self-attention mechanism, they go through a stack of convolutional neural networks (CNN) with $K$ = 3 layers inside (Figure 2). The w types of flows will be projected to a representation space with dimension $d_{model}$ = 64, and the spatial dependencies are further interrelated via the CNN stack. Previous works have shown that when predicting the flow of $v_i$, instead of measuring the whole spatial map, focusing on local dependencies is more helpful for the prediction \cite{zhang_zheng_qi}. Therefore, we also adopt the idea of local convolution, which focuses on an $a \times b$ area of interest (AoI) surrounding $v_i$. Specifically, the historical flow input $F_{input}^i \in \mathbf{R}^{a \times b \times s \times w}$ is sampled from all AoIs in $s$ historical spatial-temporal data. Similarly, when generating historical transition input $M_{input}^i \in \mathbf{R}^{a \times b \times s \times w}$, only the transitions between $v_i$ and the other nodes in the AoIs are sampled. In this work, we set \textit{a = b = 7}.
	
	The output of each layer in the CNN stack is computed as:
	
	\begin{equation}
	h_{t,p}^i = w_{t,p}^i \cdot r_t^i
	\end{equation}
	
	\noindent where $r_t^i$ $\in$ \(\mathbf{R}^{a \times b \times w}\) is a slice of $F_{input}^i$ or $M_{input}^i$, and \textit{$h_{t,p}^i$} is the convolutional result of $r_t^i$ on the \textit{p}-th channel. $w_{t,p}^i$ is the weight of the \textit{p}-th filter of convolution kernal \(\mathcal{W}_t^i\), whose filter size is $d_{model}$. All \(\mathcal{W}_{t}^i\) constitute a joint kernal \(\mathcal{W}^i\), and the final output of each layer in the CNN stack is as:
	
	\begin{equation}
	H^i = \mathcal{W}^i \ast_l R^i
	\end{equation}
	
	\noindent
	where $R^i$ is $F_{input}^i$ or $M_{input}^i$, and $H^i \in \mathbf{R}^{a \times b \times s \times d_{model}}$ is the projected spatial-temporal representation of the input data. $\ast_l$ represent the slice-wise joint convolutional operation. We employ padding with the same value for each convolutional layer to maintain the same tensor shape.
	
	\subsubsection{Periodic Shifting and Sliding-Window Sampling}
	Previous work \cite{stdn} demonstrated that the flows in periodic windows have strong similarities. As shown in Figure 4, the same periods of different days are more similar to each other than those in the previous periods on the same day. Besides, the pattern of flow will shift periodically. For example, the peak hours of traffic flow may vary from 16:30 to 18:00 on different days. Thus, we adopt sliding-window sampling to generate inputs of flow and transition from $\mathcal{F}$ and $\mathcal{M}$ to form $R$. Specifically, $R$ is the concatenation of spatial matrices from the same periods of the previous $c$ = 7 days and the previous two-time intervals of the current day (area with red boundary in Figure 4). Then, data in the time interval before the future time is used as the current data fed in the decoder stack while the remained are used as input of the encoder stack.
	
	\subsection{Positional Encoding} 
	Positional encoding is employed as the positional information is missed without the recurrent structures. Here, to encode the non-consecutive positional information, we add learned positional encodings to the output of the convolution stack. First, we represent the time information of $r_t^i$ as a one-hot vector $z_t^i \in \mathbf{R}^{7+g}$, where g is the number of time intervals in one day. We use the first seven elements of $v_t^i$ to represent the day in a week and the last g elements to represent the index of time interval in that day. The positional encoding ($PE_t^i$) of $r_t^i$ is as:
	
	\begin{equation}
	PE_t^i = \sigma(w_t^{i,1} \cdot ReLU(w_t^{i,0}z_t^i + b_t^{i,0}) + b_t^{i,1})
	\end{equation}
	
	\noindent where $w_t^{i,0}, w_t^{i_1}, b_t^{i,0}, b_t^{i,1}$ are the learned parameters, and $\sigma$ is the \textit{sigmoid} function. Then the whole positional encoding matrix $PE^i \in \mathbf{R}^{s \times (7+g)}$ is formed and summed with $H^i$ before fed in the encoder and decoder stacks. The broadcast of $PE^i$ to the same shape of $H^i$ is performed before the adding.
	
	\begin{figure}[t]
		\centering
		\includegraphics[scale = 0.35]{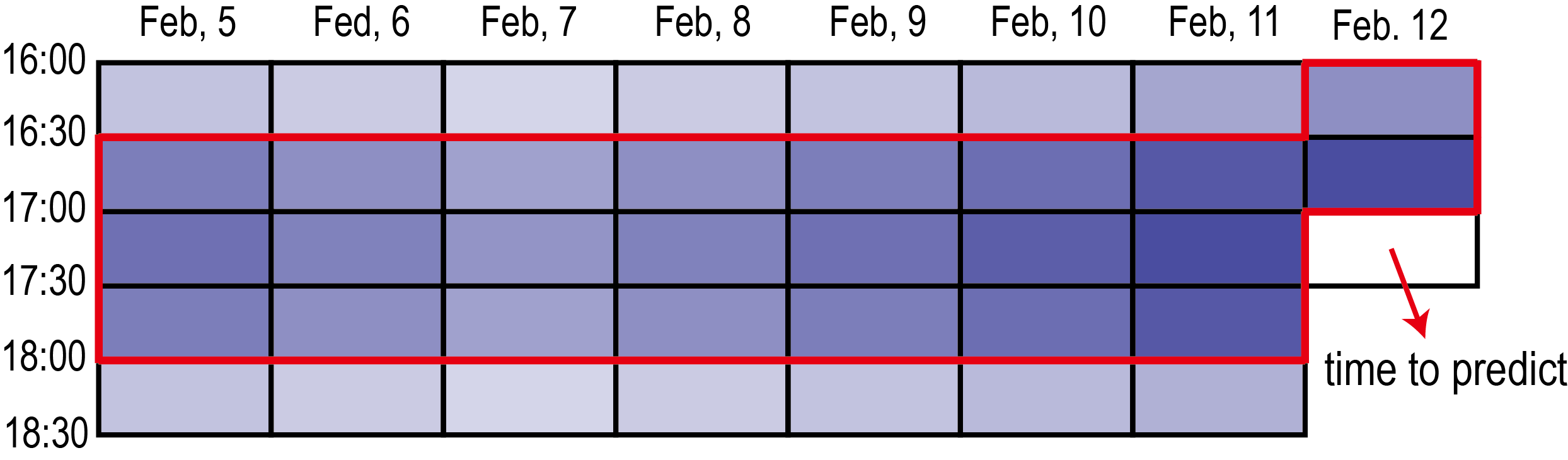} % Reduce the figure size so that it is slightly narrower than the column. Don't use precise values for figure width.This setup will avoid overfull boxes. 
		\caption{Temporal similarity. The darker a interval is the stronger its similarity to the time to predict.}
		% \label{fig3}
	\end{figure}
	
	\subsection{2-Stream Structure}
	Previous works demonstrate that transitions between nodes have significant impacts in flow prediction \cite{a98d8116a2684b17bdabc50c1e1713b3}. Therefore, ST-SAN is designed as a 2-stream framework with two spatial-temporal self-attention networks (Stream-T, Stream-F) to measure flow and transition independently. We first train the Stream-T on predicting the transitions in AoI. Here the output of the Stream-T is as:

	\begin{equation}
	\hat{y}_{transition}^i = tanh(w^i \cdot dec\_output^i + b^i)
	\end{equation}
	
	\noindent where $\hat{y}_{transition}^i \in \mathbf{R}^{a \times b \times w}$. $w^i, b^i$ are learned parameters and $dec\_output^i$ is the output from the decoder stack.
	
	Then, the trainable parameters of Stream-T will be locked and merged with Stream-F by a masked fusion mechanism to form the ST-SAN for further training. The independent training is necessary since we observe that the Stream-T will be ambiguously trained if only loss between the output and the true flow is calculated. Hence, independent training sets a more definite target for Stream-T, which enhances the measurement of transition. The experimental results also show the advantages of employing independent training.
	
	\subsection{Masked Fusion Mechanism}
	A masked fusion mechanism is proposed to merge the two streams and generate the final output. As shown in Figure 3 (b), the outputs of Stream-T ($t_0$) and Stream-F ($f_0$) are fed in a stack of $L$ = 2 hybrid convolutional layers, where its $j$-th layer's output is computed as:
	
	\begin{equation}
	\begin{aligned}
	m_j^i =\ &w_{m,j}^im_{j-1}^i + b_{m,j} \\
	e_j^i =\ &w_{f,j}^if_{j-1}^i + b_{f,j} \\
	f_j^i =\ &\sigma(m_j^i) \otimes e_j^i
	\end{aligned}
	\end{equation}
	
	\noindent where $w_{m,j}^i, w_{f,j}^i, b_{m,j}, b_{f,j}$ are the convolutional kernels and learned bias.
	The $sigmoid$ function converts the transition features to a weight mask. Then the mask is applied on the convolutional result of $f^i$ to intensify the influence of more relative nodes. To be specific, if two nodes have many transitions between, consequently their connection and mutual influences should be stronger. Here, padding is not employed in the CNN layers.
	
	After the output of the hybrid layer is flattened, the final output is then computed:
	
	\begin{equation}
	\hat{y}^i = Tanh(w_2^i \cdot ReLU(w_1^if_{flat}^i + b_1^i) + b_2^i)
	\end{equation}
	
	\noindent where $f_{flat}$ is the flattened output.
	
	The predicted outputs of all nodes \{$\hat{y}_i | \hat{y}_i \in \mathrm{R}^w and\ i \in\{1,...,N\}$\} constitute the predicted values of the whole spatial map (grid map) $Y \in \mathrm{R}^{I \times J \times w}$.
	
	\subsection{Loss}
	We use MSE loss function on both the training of Stream-T and the unified ST-SAN:

	\begin{equation}
	\mathcal{L}(\theta_t) = \frac{\sum_{i = 1}^N (\hat{y}_{transition}^i - y_{transition}^i)^2}{N \times w}
	\end{equation}
	
	\begin{equation}
	\mathcal{L}(\theta) = \frac{\sum_{i = 1}^N (\hat{y}^i - y^i)^2}{N \times w}
	\end{equation}
	
	\noindent where $y^i, y^i_{transition}$ are the ground truths of flows and AoI transitions of $v_i$ and $\theta_t$ and $\theta$ are the learnable parameters of Stream-T and ST-SAN.
	
	\begin{table*}[t]
		\centering
		\caption{Comparisons with ten baselines on Taxi-NYC, Bike-NYC, and Mobile M in flow prediction.}\smallskip
		%\resizebox{0.95\textwidth}{!}{ % If your table exceeds the column or page width, use this command to reduce it slightly
		%}
		\begin{tabular}{ l||c|c|c|c||c|c|c|c||c|c }
			\hline \hline
			\multirow{3}{*}{Model} & \multicolumn{4}{c||}{Taxi-NYC} & \multicolumn{4}{c||}{Bike-NYC} & \multicolumn{2}{c}{Mobile M} \\
			\cline{2-11}
			& \multicolumn{2}{c|}{inflow} & \multicolumn{2}{c||}{outflow} &  \multicolumn{2}{c|}{inflow} & \multicolumn{2}{c||}{outflow} & \multicolumn{2}{c}{user number} \\
			\cline{2-11}
			& RMSE & MAE & RMSE & MAE & RMSE & MAE & RMSE & MAE & RMSE & MAE \\ 
			\hline \hline
			HA & 90.19 & 50.10 & 109.36 & 65.91 & 30.25 & 20.35 & 29.63 & 19.96 & 421.39 & 273.18 \\
			ARIMA & 33.54 & 18.62 & 40.70 & 23.61 & 17.14 & 10.83 & 18.03 & 11.28 & 194.92 & 150.95 \\
			VAR & 48.04 & 23.21 & 128.67 & 29.84 & 27.37 & 14.29 & 27.67 & 15.09 & 254.37 & 157.71 \\
			MLP & 27.13 & 16.91 & 32.93 & 20.80 & 25.77 & 32.57 & 15.92 & 19.85 & 130.01 & 106.44 \\
			LSTM & 24.35 & 15.07 & 30.41 & 19.18 & 24.79 & 32.06 & 15.61 & 20.62 & 111.70 & 93.80 \\
			GRU & 24.37 & 15.17 & 30.25 & 19.14 & 24.62 & 31.37 & 15.22 & 19.77 & 114.23 & 93.89 \\
			ConvLSTM & 22.25 & 14.13 & 27.39 & 17.38 & 9.71 & 7.07 & 11.09 & 7.78 & 85.97 & 67.12 \\
			ST-ResNet & 20.34 & 12.90 & 25.54 & 16.21 & 9.32 & 6.79 & 10.45 & 7.33 & 74.30 & 55.03\\
			DMVST-Net & 18.99 & 12.24 & 24.07 & 15.39 & 8.95 & 6.52 & 9.75 & 6.84 & 68.09 & 50.50\\
			STDN & 17.91 & 11.37 & 23.47 & 14.89 & 8.58 & 6.25 & 9.44 & 6.62 & 62.59 & 43.22\\
			ST-SAN & \textbf{16.39} & \textbf{10.63} & \textbf{22.94} & \textbf{13.48} & \textbf{7.82} & \textbf{5.68} & \textbf{9.02} & \textbf{6.17} & \textbf{57.13} & \textbf{40.20}\\
			\hline \hline
		\end{tabular}
	\end{table*}
	
	\section{Experiment}
	\subsection{Datasets}
	We evaluate our model on three real-world datasets -- Taxi-NYC, Bike-NYC, and Mobile M. Their details are showed in Table 2.
	
	\begin{itemize}
		\item \textbf{Taxi-NYC and Bike-NYC:} Taxi-NYC and Bike-NYC both contain 60 days of trip records. Each record includes the locations and times of the start and the end of a trip. We use the first 40 days as training data, and the remained 20 days as testing data.
		\item \textbf{Mobile M}: Mobile M includes 158,742,004 service records that contain the approximate locations of mobile phone users during the service periods. The whole 92-day dataset is split to 60 and 32 days for training and testing.
	\end{itemize}
	
	\subsection{Evaluation Metric \& Baselines} We measure the performance of different methods by two widely adopted metrics: (1) Rooted Mean Square Error (RMSE); (2) Mean Absolute Error (MAE).
	
	\begin{table}[t]
		\centering
		\caption{Details of the evaluated datasets}\smallskip
		%\resizebox{0.95\textwidth}{!}{ % If your table exceeds the column or page width, use this command to reduce it slightly
		%}
		\begin{tabular}{ c|c|c|c }
			\hline \hline
			Datasets & Taxi-NYC & Bike-NYC & Modile M \\
			\hline
			Grid map size & $16 \times 12$ & $14 \times 8$ & $16 \times 16$ \\
			\hline
			Time interval & 30 mins & 30 mins & 15 mins \\
			\hline
			\multirow{2}{*}{Time Span} & 1/1/2016 - & 8/1/2016 - & 10/1/2018 - \\
			& 2/29/2016 & 9/29/2016 & 12/31/2018\\
			\hline
			Total records & 22,437,649 & 9,194,087 & 158,742,004 \\
			\hline \hline
		\end{tabular}
	\end{table}
	
	\subsubsection{Baselines}
	\begin{itemize}
		\item \textbf{HA}: Historical average.
		\item \textbf{ARIMA}: Auto-regressive integrated moving average model.
		\item \textbf{VAR}: Vector auto-regressive model.
		\item \textbf{MLP}: Multi-layer perceptron.
		\item \textbf{LSTM}: Long-Short-Term-Memory \cite{doi:10.1162/neco.1997.9.8.1735}.
		\item \textbf{GRU}: Gated-Recurrent-Unit network \cite{DBLP:journals/corr/ChungGCB14}.
		\item \textbf{ConvLSTM}: Convolutional LSTM \cite{NIPS2015_5955}.
		\item \textbf{ST-ResNet}: Spatial-Temporal Residual Convolutional Network \cite{Zhang:2017:DSR:3298239.3298479}.
		\item \textbf{DMVST-Net}: Deep Multi-View Spatial-Temporal Network\cite{DBLP:journals/corr/abs-1802-08714}.
		\item \textbf{STDN}: Spatial-Temporal Dynamic Network \cite{stdn}.
	\end{itemize}
	
	\subsection{Preprocessing} The grid sizes of Taxi-NYC, Bike-NYC, and Mobile M are $16 \times 12$, $14 \times 8$, and $16 \times 16$ respectively. The length of the time interval is set as 30 minutes and 15 minutes, whereas the number of time interval in every day is 48 and 96. We randomly select 20\% of data of training dataset for validation and the remained for training. We use Min-Max normalization to convert all traffic flow data to scale of [0, 1], and convert them back during the evaluation. We also filter out all regions with real flow volume less than ten in the evaluation, which is a common criterion used in flow prediction research area \cite{Zhang:2017:DSR:3298239.3298479}. 
	
	\subsection{Hyperparameters}
	In Taxi-NYC and Bike-NYC, $w$ = 2 types of flow -- inflow and outflow, are processed. In Mobile M, only user number of each area is considered ($w$ = 1). We set threshold m = 2 to filter out long-span transitions. The stack of convolutional layers contains $K$ = 3 layers of CNN, each of which includes $d_{model}$ = 64 filters with kernel size = \textit{$3 \times 3$}. We set the dimension of Feed-Forward layer to 128 and the number of attention head to 8. The dropout rate is 0.1, and the epsilon offset in layer normalization is \textit{-1e6}. 
	
	\subsection{Optimizer}
	We used the Adam optimizer \cite{adam} with $\beta_1$ = 0.9, $\beta_2$ = 0.98 and $\epsilon = 10^{-9}$. We adopted warm-up to adjust the learning rate:
	
	\begin{equation}
	lr = d_{model}^{-0.5}\cdot min(steps^{-0.5},steps\cdot wu\_steps^{-1.5})
	\end{equation}
	
	\noindent where $wu\_steps$ = 4000.
	
	\subsection{Results}
	We evaluated our methods and ten baselines on all three datasets and obtained the average results of each method after ten executions. Table 1 demonstrates the results of RMSE and MAE.
	
	Noticeably, traditional statistic time-series prediction methods (HA, ARIMA, and VAR) are significantly less effective. It exposes the weakness of methods of exclusively considering the relation of historical statistic values and ignoring the complicated spatial-temporal dependency. For MLP, it barely learned the linear mapping from historical data to the predicted results, the spatial-temporal dependencies are insufficiently measured. LSTM and GRU achieved non-trivial improvement compared to MLP and traditional time-series methods given their effectiveness on modeling temporal dependency. Nonetheless, without a sophisticated mechanism to integrate spatial dependencies, their performance failed to improve further.
	
	Deep-learning based methods showed their advantage of capturing complicated spatial-temporal dependencies. As shown in the comparison result, ST-SAN has outperformed the other deep learning frameworks.
	For ST-ResNet, despite it employs deep residual networks to capture spatial-temporal dependencies, the convolutional results are linearly merged, which overlooks the distinctive impacts of short-term and long-term temporal dependencies. ConvLSTM, DMVST-Net, and STDN showed the remarkable capability of modeling both the spatial and temporal dependencies. However, the LSTM employed limits their efficiencies on reaching long-term temporal dependencies. Besides, independent modeling of spatial and temporal dependencies also limits their capacity of capturing complicate spatial-temporal correlations. ST-SAN shows significant improvement compared to previous deep learning methods. In details, taking the prediction on Taxi-NYC data as an example, the RMSE is reduced by 9\% for inflow prediction and 4\% for outflow prediction.
	
	\subsection{Model Variants}
	\subsubsection{Evaluation on the Effectiveness of Spatial-Temporal Self-Attention Mechanism} In this section, we empirically demonstrate the effectiveness of the spatial-temporal self-attention mechanism. There are three variants of the self-attention networks:
	
	\begin{table}[t]
		\centering
		\caption{Evaluation of variants of ST-SAN on Taxi-NYC.}\smallskip
		%\resizebox{0.95\textwidth}{!}{ % If your table exceeds the column or page width, use this command to reduce it slightly
		%}
		\begin{tabular}{ l||c|c }
			\hline \hline
			\multirow{2}{*}{Variants} & \multicolumn{2}{c}{RMSE/MAE} \\
			\cline{2-3}
			& inflow & outflow \\
			\hline
			SAN & 22.38/13.98 & 28.17/16.44 \\
			ST-SAN-S & 19.38/12.98 & 24.97/16.44 \\
			ST-SAN-D & 16.73/10.91 & 23.34/14.07 \\
			ST-SAN-D IT & \textbf{16.39/10.63} & \textbf{22.94/13.48} \\
			\hline \hline
		\end{tabular}
	\end{table}
	
	\begin{itemize}
		\item \textbf{SAN}: Original self-attention network. The spatial maps are embedded into vectors by fully connected layers. Except for the input and output layers, SAN is identical to the Transformer.
		\item \textbf{ST-SAN-S}: Single-stream ST-SAN employing spatial-temporal self-attention network.
		\item \textbf{ST-SAN-D}: Dual-stream (2-stream) ST-SAN without independent training on Stream-T.
	\end{itemize}
	
	\noindent As shown in Table 3, ST-SAN-D outperforms other variants based on RMSE and MAE. SAN obtains poor performance as it merely employs the structure of Transformer ignoring the complicated spatial dependencies. ST-SAN-S applies the spatial-temporal self-attention mechanism, but the transition information between nodes is missed, which leads to the uniform measurement of influences of other nodes and overlooks their dynamic dependencies.
	
	\subsubsection{Evaluation on Effectiveness of Independent Training}
	To demonstrate the effectiveness of independent training on Stream-T, we evaluate the performance of 2 variants: 
	
	\begin{itemize}
		\item \textbf{ST-SAN-D}
		\item \textbf{ST-SAN-D IT}: ST-SAN with independent training on Stream-T.
	\end{itemize}
	
	\noindent The results demonstrated in Table 3 show that ST-SAN-D MT achieves reasonable improvement compared to other variants. As mentioned above, if ST-SAN is only trained toward predicting flow, the target of Stream-T is ambiguous. Therefore, independent training of Stream-T reduces the ambiguity, leading to more accurate modeling of connectivity between nodes. Consequently, the final training on flow prediction task benefits from the pre-training.
	
	\section{Conclusion and Future Work}
	In this work, we present the spatial-temporal self-attention network. We introduce a spatial-temporal self-attention mechanism that simultaneously captures spatial and temporal dependencies while measuring long-term dependencies more efficiently. In addition, we proposed an independent training scheme to enhance the network's ability to measure the connectivities of nodes. Experiment results demonstrate the significant improvement achieved by ST-SAN. In future work, we will focus on improving the performance of outflow prediction. During the experiment, we observed that ST-SAN achieved much fewer improvement on outflow prediction compared to inflow prediction. To find out the reason is one of the main tasks of our future works.
	
	\bibliography{references.bib}
	\bibliographystyle{aaai}
\end{document}